\documentclass[10pt,a4paper]{article}

\usepackage[a4paper,margin=0.72in]{geometry}
\usepackage[T1]{fontenc}
\usepackage[utf8]{inputenc}
\usepackage{lmodern}
\usepackage{microtype}
\usepackage{amsmath,amssymb,mathtools}
\usepackage{booktabs}
\usepackage{array}
\usepackage{tabularx}
\usepackage{multirow}
\usepackage{makecell}
\usepackage{siunitx}
\usepackage{xcolor}
\usepackage{colortbl}
\usepackage{tikz}
\usepackage{pgfplots}
\usepackage{pgfplotstable}
\usepackage[most]{tcolorbox}
\usepackage{enumitem}
\usepackage{graphicx}
\usepackage{float}
\usepackage{caption}
\usepackage{subcaption}
\usepackage{hyperref}
\usepackage{cleveref}
\usepackage{fancyhdr}
\usepackage{titlesec}
\usepackage{ragged2e}
\usetikzlibrary{positioning}
\usepackage{lastpage}
\usepackage{url}

\pgfplotsset{compat=1.18}

\definecolor{OneBitNavy}{HTML}{12263A}
\definecolor{OneBitBlue}{HTML}{1769AA}
\definecolor{OneBitCyan}{HTML}{00A6A6}
\definecolor{OneBitGreen}{HTML}{2E8B57}
\definecolor{OneBitGold}{HTML}{E6A817}
\definecolor{OneBitOrange}{HTML}{E67E22}
\definecolor{OneBitRed}{HTML}{C94C4C}
\definecolor{SoftBlue}{HTML}{EAF4FB}
\definecolor{SoftCyan}{HTML}{E9F8F7}
\definecolor{SoftGreen}{HTML}{EDF8F1}
\definecolor{SoftGold}{HTML}{FFF7DF}
\definecolor{SoftRed}{HTML}{FCEEEE}
\definecolor{SoftGray}{HTML}{F4F6F8}
\definecolor{MidGray}{HTML}{667085}
\definecolor{DarkGray}{HTML}{273444}

\hypersetup{
    colorlinks=true,
    linkcolor=OneBitBlue,
    citecolor=OneBitBlue,
    urlcolor=OneBitBlue,
    pdftitle={Capability-Stratified Degradation in Ternary Language Models},
    pdfauthor={OneBit AI}
}

\titleformat{\section}
  {\Large\bfseries\color{OneBitNavy}}
  {\thesection}{0.55em}{}
  [\vspace{-3pt}\color{OneBitCyan}\rule{\textwidth}{1.2pt}]

\titleformat{\subsection}
  {\large\bfseries\color{OneBitBlue}}
  {\thesubsection}{0.5em}{}

\titleformat{\subsubsection}
  {\normalsize\bfseries\color{DarkGray}}
  {\thesubsubsection}{0.5em}{}

\renewcommand{\headrulewidth}{0.3pt}
\renewcommand{\headrule}{\hbox to\headwidth{\color{SoftGray}\leaders\hrule height \headrulewidth\hfill}}

\tcbset{
    enhanced,
    boxrule=0pt,
    arc=3pt,
    left=7pt,right=7pt,top=7pt,bottom=7pt,
    coltitle=OneBitNavy,
}

\newtcolorbox{keybox}[1][]{
    colback=SoftBlue,
    borderline west={3pt}{0pt}{OneBitBlue},
    #1
}

\newtcolorbox{insightbox}[1][]{
    colback=SoftCyan,
    borderline west={3pt}{0pt}{OneBitCyan},
    #1
}

\newtcolorbox{businessbox}[1][]{
    colback=SoftGold,
    borderline west={3pt}{0pt}{OneBitGold},
    #1
}

\newtcolorbox{caveatbox}[1][]{
    colback=SoftGray,
    borderline west={3pt}{0pt}{MidGray},
    #1
}

\newcommand{\good}[1]{\textcolor{OneBitGreen}{\textbf{#1}}}

\newcommand{\bad}[1]{\textcolor{OneBitRed}{\textbf{#1}}}
\newcommand{\cloe}{\textsc{Cloe}}

\newcolumntype{Y}{>{\RaggedRight\arraybackslash}X}
\newcolumntype{C}{>{\Centering\arraybackslash}X}

\begin{document}

% ------------------------- Title -----------------------------
\begin{center}
    {\fontsize{22}{26}\selectfont\bfseries\color{OneBitNavy}
    Capability-Stratified Degradation in Ternary Language Models}\\[5pt]
    {\fontsize{15}{18}\selectfont\color{OneBitBlue}
    CLOE V1.1 - OneBit AI}\\[10pt]

    \begin{tcolorbox}[
        width=0.94\textwidth,
        colback=SoftCyan,
        borderline west={4pt}{0pt}{OneBitCyan},
        halign=center
    ]
    \large
    \textbf{Qwen3.5-0.8B $\rightarrow$ QAT $\rightarrow$ ternary Cloe}

    \vspace{3pt}
    \normalsize
    A capability, representation, learnability, and deployment study of a
    \SI{752}{M}-parameter pretrained language model under ternary conversion.
    \end{tcolorbox}

    \vspace{4pt}
    {\small\color{MidGray}
    Anirudh Malik \quad|\quad Poojith Devan \quad|\quad M Sparsh Mehra}
\end{center}

\vspace{4pt}

% ---------------------- Executive takeaway ------------------
\begin{keybox}
\textbf{\large Executive takeaway.}
The central result is not that ternarisation produces a uniformly weaker language
model. Instead, degradation is \textbf{capability-stratified}: specialist factual
knowledge and language-modelling quality degrade sharply, while several forms of
commonsense judgement, representational capacity, and downstream task learnability
remain measurable. Across ten benchmarks where Cloe demonstrably discriminates
between candidate answers, it retains a mean \textbf{77.1\%} of the full-precision
teacher. After task-specific fine-tuning, it reaches \textbf{95.6\%} of an identically
fine-tuned teacher on SST-2 and \textbf{79.4\%} on XSum. The engineering study further
shows a \textbf{634.7 MB packed artifact} and higher prefill/decode throughput than
the measured latent representation, while runtime memory remains a separate
constraint.
\end{keybox}

\begin{abstract}
Extreme low-bit inference offers a route toward smaller model artifacts and more
constrained deployment. Ternary language models restrict weights to
$\{-1,0,+1\}$, approaching the information-theoretic limit of
$\log_2 3 \approx 1.585$ bits per weight. The practical question for an existing
pretrained model, however, is not simply whether the weights can be quantised, but
\emph{which capabilities survive the transformation} and whether the resulting
model remains useful as a substrate for downstream adaptation.

We study this question by converting Qwen3.5-0.8B, a \SI{752}{M}-parameter
pretrained model, to ternary weights using 72.4M tokens of quantisation-aware training (QAT).
The resulting model, Cloe, is evaluated across 29 benchmarks and through
representation-level diagnostics, matched downstream fine-tuning, and deployment
measurements. The evidence shows a non-uniform degradation pattern. A linear probe
recovers 43.76\% of MMLU answers from the full-precision teacher's final-layer
representations but only 26.19\% from Cloe, close to the 25\% chance level,
supporting the interpretation that specialist factual information is substantially
lost rather than merely hidden behind an impaired output channel. At the same time,
Cloe retains measurable performance on ten tasks, averaging 77.1\% of teacher
performance on that interpretable subset. More importantly for application-specific
deployment, fine-tuning raises Cloe from 61.9\% to 89.8\% on SST-2, corresponding
to 95.6\% of the matched fine-tuned teacher, while XSum reaches 79.4\% teacher
retention. Because the available QAT budget was limited to 72.4M tokens, we attribute the observed degradation conservatively to a combination of quantization-induced information loss and incomplete recovery, rather than to ternarisation alone

We also identify an evaluation pitfall: standard answer-letter scoring initially
reported 22.97\% MMLU accuracy because Cloe emitted ``A'' on 98.6\% of questions.
Continuation scoring and majority-class guards are therefore used throughout the
interpretable evaluation. The study consequently argues for a capability-aware
view of extreme quantisation: a ternary conversion may be unsuitable as a
drop-in replacement for a general-purpose pretrained model, yet remain valuable as
a compact, adaptable substrate for application-specific models.
\end{abstract}


% ============================================================
\section{Introduction}

The deployment economics of language models are increasingly shaped by the cost of
moving, storing, and executing model weights. Reducing numerical precision is one of
the most direct ways to reduce the information carried by a parameter tensor.
Ternary models take this idea to an extreme by constraining weights to
$\{-1,0,+1\}$, for which the theoretical information content is
$\log_2 3 \approx 1.585$ bits per weight \cite{bitnetb158}.

The research landscape has already established that models can be trained natively
under such low-bit constraints ~\cite{bitnet,bitnetb158}. The more operationally difficult question is what
happens when an \emph{already pretrained} model is pushed into this regime. This
distinction matters for practitioners because pretrained models contain a large
amount of accumulated factual, linguistic, and task-relevant structure. If that
structure cannot survive conversion, a smaller checkpoint is not necessarily a
more useful model. 

This work therefore takes a deliberately diagnostic perspective. Rather than reducing
the outcome to a single aggregate benchmark score, we ask four connected questions:

\begin{enumerate}[leftmargin=1.5em,itemsep=1pt]
    \item \textbf{Capability:} Which behaviours survive ternarisation?
    \item \textbf{Representation:} Is poor task performance caused by representational
    collapse, or by loss of specific stored information?
    \item \textbf{Learnability:} Can the compressed model still acquire a new task
    through supervised adaptation?
    \item \textbf{Deployment:} What does the packed representation actually buy in
    storage and inference measurements?
\end{enumerate}

This produces a more useful engineering picture than a single ``accuracy after
quantisation'' number.

\begin{businessbox}
\textbf{Why this matters for an AI product.}
A general-purpose model and an application-specific model have different requirements.
A general-purpose model must preserve broad knowledge. An application-specific model
can instead be judged by whether it can efficiently represent and execute the
capability required by a particular product. The distinction between \emph{knowledge
retention} and \emph{task learnability} is therefore commercially relevant: it
determines whether extreme compression should be viewed only as destructive
compression or also as a possible route toward compact specialised models.
\end{businessbox}

% ============================================================
\section{Research Positioning and Related Work}

Prior work on ternary language models can be broadly separated into native low-bit
training and conversion of pretrained models ~\cite{gptq,catq,twla,ptqtp,ayot}. BitNet b1.58 established the
feasibility of training language models with ternary weights. More recent approaches
such as CAT-Q, PTQTP, ScaleQ-1.58, and Ternary Mamba investigate how pretrained
models can be converted or adapted into low-bit regimes. Falcon-Edge represents
another direction: native 1.58-bit pretrained models rather than conversion of an
existing full-precision checkpoint \cite{ternarymamba,falconedge}.

The distinction is important for positioning this study. The contribution here is
\emph{not} claimed to be a new state-of-the-art ternary quantiser. Instead, the work
uses QAT to create a controlled ternary conversion and then asks what information,
behaviour, and adaptability survive.

\begin{table}[H]
\centering
\caption{Positioning against major directions represented in the project literature.}
\label{tab:related}
\small
\begin{tabularx}{\textwidth}{p{2.1cm}p{2.0cm}p{2.4cm}Y}
\toprule
\textbf{Work / direction} & \textbf{Model regime} & \textbf{Core approach} &
\textbf{Primary emphasis} \\
\midrule
BitNet b1.58 & Native ternary & Train directly at low precision &
Demonstrates viability of ternary training and inference. \\

CAT-Q & Pretrained conversion & Calibration / post-training conversion &
Demonstrates that pretrained models can be brought toward ternary operation. \\

PTQTP & Pretrained conversion & Dual trit-plane representation &
Improves the representation of ternary weights in PTQ settings. \\

ScaleQ-1.58 & Pretrained conversion & Calibration with additional generated signals &
Explores stronger post-training ternary conversion. \\

Ternary Mamba & Pretrained conversion & Grouped quantisation + distillation / QAT &
Closest precedent in spirit for converting an existing model and reporting retention. \\

Falcon-Edge & Native low-bit & Native 1.58-bit pretraining &
Shows the alternative strategy of avoiding conversion altogether. \\

\rowcolor{SoftBlue}
\textbf{This work} & \textbf{Pretrained conversion} &
\textbf{QAT ternarisation + capability diagnostics} &
\textbf{Characterises what is lost, what survives, and whether downstream
learnability remains after conversion.} \\
\bottomrule
\end{tabularx}
\end{table}

The comparison in \cref{tab:related} defines the paper's central distinction. A
conversion method is normally evaluated by asking whether the resulting model
retains benchmark performance. Here we additionally ask whether a failed benchmark
result reflects missing information, an impaired output channel, representational
collapse, or a failure that can be repaired through task-specific training. This
makes the study closer to a \emph{capability characterization} than a conventional
quantisation leaderboard entry.

% ============================================================
\section{Model and Ternary Conversion}

\subsection{Base model}

Cloe is derived from Qwen3.5-0.8B, with approximately \SI{752}{M} parameters \cite{qwen35}.
The model contains 24 transformer layers, with 18 linear-attention layers and
6 full-attention layers. Its hidden size is 1024 and its head dimension is 256.
The vocabulary contains 248,320 entries, with tied embeddings and language-model
head.

\begin{table}[H]
\centering
\caption{Core configuration of the Cloe conversion.}
\label{tab:model}
\small
\begin{tabularx}{0.92\textwidth}{p{0.48\textwidth}C}
\toprule
\textbf{Property} & \textbf{Value} \\
\midrule
Base model & Qwen3.5-0.8B \\
Parameters & \SI{752}{M} \\
Transformer layers & 24 \\
Attention structure & 18 linear-attention + 6 full-attention \\
Hidden size & 1024 \\
Head dimension & 256 \\
Vocabulary & 248,320 \\
Ternary linear layers & \good{186 / 186} \\
Distinct weight values per ternary layer & \good{Exactly 3} \\
Measured ternary entropy & \good{1.5821 bits/weight} \\
Ideal ternary entropy & 1.5850 bits/weight \\
Cumulative QAT tokens & 72.4M \\
\bottomrule
\end{tabularx}
\end{table}

The values in \cref{tab:model} establish that the conversion is not merely a
quantisation-inspired approximation. Every targeted linear layer is verified to
contain exactly three distinct weight values, and the measured entropy of
1.5821 bits/weight is very close to the theoretical ternary limit. This distinction
is important because a model can otherwise be described as ``ternary'' while
retaining continuous-valued weights in parts of its computation.

\subsection{Ternary parameterisation}

Each targeted linear layer uses abs-mean scaling and a straight-through estimator \cite{ste}.
The effective quantisation can be represented as

\begin{align}
s &= \operatorname{mean}(|W|),\\
Q(W) &= \operatorname{round}\left(
    \operatorname{clamp}\left(\frac{W}{s},-1,1\right)
    \right)s,\\
W_{\mathrm{eff}}
&= W + \lambda\left(Q(W)-W\right)_{\mathrm{detach}}.
\end{align}

At inference, $\lambda=1$. An RMSNorm is applied to activations before the
low-precision matrix multiplication \cite{rmsnorm158}.

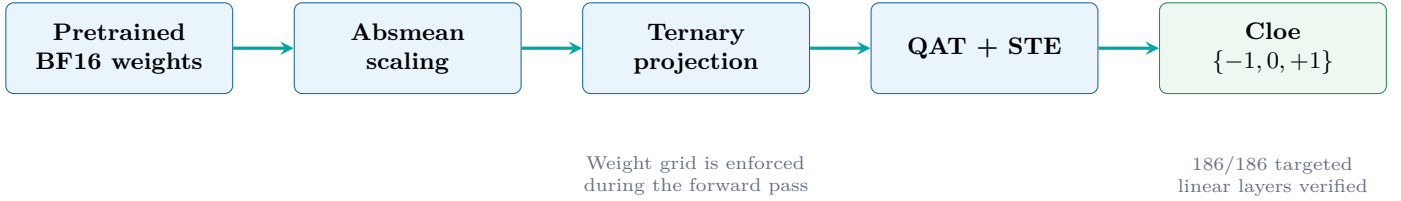
\begin{figure}[H]
\centering
\begin{tikzpicture}[
    node distance=7mm and 8mm,
    box/.style={rounded corners=3pt, draw=OneBitBlue, fill=SoftBlue,
                minimum width=30mm, minimum height=12mm, align=center,
                font=\small\bfseries},
    arrow/.style={->,>=stealth,very thick,draw=OneBitCyan}
]
\node[box] (fp) {Pretrained\\BF16 weights};
\node[box,right=of fp] (scale) {Absmean\\scaling};
\node[box,right=of scale] (quant) {Ternary\\projection};
\node[box,right=of quant] (qat) {QAT + STE};
\node[box,right=of qat,fill=SoftGreen,draw=OneBitGreen] (cloe)
    {\cloe\\$\{-1,0,+1\}$};

\draw[arrow] (fp)--(scale);
\draw[arrow] (scale)--(quant);
\draw[arrow] (quant)--(qat);
\draw[arrow] (qat)--(cloe);

\node[below=7mm of quant,align=center,text width=32mm,font=\scriptsize\color{MidGray}]
    {Weight grid is enforced during the forward pass};
\node[below=7mm of cloe,align=center,text width=32mm,font=\scriptsize\color{MidGray}]
    {186/186 targeted linear layers verified};
\end{tikzpicture}
\caption{Overview of the conversion pipeline used to obtain Cloe.}
\label{fig:pipeline}
\end{figure}

\subsection{Verification of the ternary execution path}

A potential implementation failure in low-bit systems is that a checkpoint may
contain a quantisation mechanism while the actual forward pass silently uses
latent full-precision weights. The project therefore verifies the loaded
quantisation state across all 186 targeted layers. The reconstructed weights differ
from the exact ternary grid only by floating-point rounding at a relative scale of
approximately $1.8\times10^{-6}$. Setting $\lambda=0$ and running the latent
weights produces gibberish, providing a direct sanity check that the evaluated
model depends on the ternary path.

\begin{insightbox}
\textbf{Interpretation.}
This verification closes a common reproducibility gap in quantisation experiments:
the reported behaviour is attributable to the quantised computation rather than
an accidentally retained full-precision path.
\end{insightbox}

\subsection{Important accounting distinction}

The phrase ``1.58-bit model'' refers to the ternary weight representation itself,
not necessarily to the complete checkpoint. The 248,320-token embedding table
remains BF16 and is tied to the language-model head. Consequently, the overall
checkpoint is reported in the project as approximately 6.47 bits/parameter on disk,
rather than a literal 1.58 bits/parameter for the entire model.

This distinction is preserved throughout the paper because it separates a
\emph{weight-representation result} from a \emph{whole-model storage result}.

% ============================================================
\section{Evaluation Framework}

The evaluation is designed around a simple principle: a low-bit model should not
be declared capable merely because a raw benchmark number is numerically high.
Conversely, a low number should not automatically be interpreted as proof that the
underlying capability has disappeared.

\begin{table}[H]
\centering
\caption{Evaluation instruments and the question each instrument answers.}
\label{tab:evaluation}
\small
\begin{tabularx}{\textwidth}{p{3.0cm}p{4.0cm}Y}
\toprule
\textbf{Instrument} & \textbf{What it measures} & \textbf{Why it is included} \\
\midrule
Continuation scoring & Candidate-answer likelihood &
Avoids relying on an answer-letter output channel. \\

Majority baseline & Trivial class-selection performance &
Detects cases where raw accuracy is inflated by label imbalance. \\

Balanced accuracy & Class-balanced discrimination &
Prevents constant predictions from being interpreted as capability. \\

Linear probe & Recoverable task information in hidden states &
Separates inaccessible/removed information from output-head problems. \\

Effective rank & Diversity of representation directions &
Tests whether quantisation causes broad representational collapse. \\

Matched fine-tuning & Downstream task learnability &
Tests whether useful task structure can still be acquired after conversion. \\

Perplexity & Language-modelling quality &
Measures the cost to token-level modelling rather than task classification. \\
\bottomrule
\end{tabularx}
\end{table}

The evaluation therefore moves from \emph{behaviour} to \emph{representation} to
\emph{learning}. This progression is important: a benchmark tells us that a model
failed a task, while a representation probe and a matched fine-tuning experiment
help explain whether the failure reflects lost information or a loss of adaptability.

\subsection{The answer-letter failure}

The original multiple-choice evaluation produced 22.97\% MMLU accuracy, which
initially suggested catastrophic failure \cite{mmlu}. Inspection of the generated answer
distribution revealed that Cloe emitted ``A'' for 13,852 of 14,042 questions,
or 98.6\% of the evaluation set.

\begin{table}[H]
\centering
\caption{Answer-letter distribution that exposed the degenerate MMLU output channel.}
\label{tab:letter}
\small
\begin{tabular}{lrrrr}
\toprule
\textbf{Model} & \textbf{A} & \textbf{B} & \textbf{C} & \textbf{D}\\
\midrule
Cloe, pre-SFT & \cellcolor{SoftRed}\bad{13,852} & 113 & 14 & 63\\
Cloe, post-SFT & 6,801 & 2,505 & 1,248 & 3,488\\
BF16 teacher & 301 & 2,914 & 2,524 & \cellcolor{SoftGold}\textbf{8,303}\\
\bottomrule
\end{tabular}
\end{table}

The implication is methodological rather than merely negative. A score produced by
answer-letter elicitation is only meaningful if the model distributes its outputs
in a way that permits genuine discrimination. The project therefore replaces this
instrument with length-normalised continuation log-probability over the candidate
answer text. This also makes the reported benchmark results more comparable to the
ternary literature represented by BitNet, Ternary Mamba, and CAT-Q \cite{bitnet,bitnetb158,catq,ternarymamba}.

\subsection{False-positive guards}

A task is treated as a demonstrated capability only when it exceeds both a
chance/majority threshold and a balanced-accuracy threshold by a two-standard-error
margin:

\begin{align}
\mathrm{acc}_{norm} &>
\max(\mathrm{chance},\mathrm{majority}) + 2\,SE,\\
\mathrm{balanced} &>
\mathrm{chance}+2\,SE.
\end{align}

This guard is important because CoLA, MRPC, and BoolQ produced superficially
strong raw accuracies while emitting a constant label. Balanced accuracy exposed
these as degenerate outcomes.

\begin{caveatbox}
\textbf{How failures are treated in this paper.}
Invalid or degenerate measurements are reported where necessary to establish
measurement validity, but they are not treated as headline capability results.
The main results use the guarded evaluation protocol.
\end{caveatbox}

% ============================================================
\section{Capability Retention}

The central behavioural result is that ternarisation does not degrade every
capability equally. On the subset of ten benchmarks for which Cloe demonstrably
discriminates between candidate answers, mean performance is 77.1\% of the
full-precision teacher.

\subsection{Teacher--Cloe comparison}

To obtain a capability-retention estimate that is not confounded by tasks on which the evaluation fails to demonstrate meaningful discrimination, we first apply the discrimination guard. Only tasks satisfying both the majority-class and chance-level balanced-accuracy criteria are included in the primary retention aggregate. This procedure identifies ten tasks in the present evaluation. The reported mean retention therefore summarizes these ten discriminative tasks only; the remaining tasks are retained in the analysis and reported separately in table \ref{tab:boundaries}.

\begin{table}[H]
\centering
\caption{Benchmark performance and capability retention relative to the full-precision teacher. ``Bar'' denotes the baseline ternary model, while ``Over bar'' reports the absolute Cloe improvement over that baseline. Tasks shown are those satisfying both the majority-baseline and chance-level balanced-accuracy criteria. Retention is computed relative to the full-precision teacher.}
\label{tab:retained}
\small
\setlength{\tabcolsep}{4pt}
\begin{tabularx}{\textwidth}{
    l
    X
    >{\centering\arraybackslash}p{1.35cm}
    >{\centering\arraybackslash}p{1.35cm}
    >{\centering\arraybackslash}p{1.35cm}
    >{\centering\arraybackslash}p{1.15cm}
    >{\centering\arraybackslash}p{1.25cm}
}
\toprule
\textbf{Task} &
\textbf{Capability} &
\textbf{Cloe} &
\textbf{Teacher} &
\textbf{Retention} &
\textbf{Bar} &
\textbf{Over bar} \\
\midrule

\texttt{race}
& Reading
& \textbf{34.0\%}
& 38.3\%
& \cellcolor{SoftGreen}\good{89\%}
& 28.0\%
& +6.0 \\

\texttt{piqa}
& Physical commonsense
& \textbf{60.7\%}
& 69.0\%
& \cellcolor{SoftGreen}\good{88\%}
& 50.7\%
& +10.0 \\

\texttt{commonsenseqa}
& Commonsense
& \textbf{34.0\%}
& 39.7\%
& \cellcolor{SoftGreen}\good{86\%}
& 23.7\%
& +10.3 \\

\texttt{qasc}
& Knowledge-easy
& \textbf{31.3\%}
& 37.3\%
& \cellcolor{SoftGreen}\good{84\%}
& 15.0\%
& +16.3 \\

\texttt{sciq}
& Knowledge-easy
& \textbf{54.7\%}
& 66.0\%
& \cellcolor{SoftGreen}\good{83\%}
& 27.0\%
& +27.7 \\

\texttt{hellaswag}
& Commonsense
& \textbf{37.3\%}
& 47.3\%
& \cellcolor{SoftGreen}\good{79\%}
& 28.0\%
& +9.3 \\

\texttt{swag}
& Commonsense
& \textbf{42.0\%}
& 59.3\%
& \cellcolor{SoftGreen}\good{71\%}
& 28.0\%
& +14.0 \\

\texttt{arc\_easy}
& Knowledge-easy
& \textbf{38.7\%}
& 55.7\%
& \cellcolor{SoftGreen}\good{69\%}
& 30.3\%
& +8.3 \\

\texttt{sst2}
& Sentiment
& \textbf{59.3\%}
& 88.0\%
& \cellcolor{SoftGreen}\good{67\%}
& 51.7\%
& +7.7 \\

\texttt{ag\_news}
& Topic
& \textbf{36.0\%}
& 66.0\%
& \cellcolor{SoftGreen}\good{55\%}
& 29.3\%
& +6.7 \\

\midrule
\rowcolor{SoftBlue}
\textbf{Mean}
& \textbf{Ten-task benchmark mean}
& \textbf{43.8\%}
& \textbf{56.3\%}
& \cellcolor{SoftBlue}\textbf{77.1\%}
& \textbf{31.2\%}
& \textbf{+11.6} \\

\bottomrule
\end{tabularx}
\end{table}

The most important feature of \cref{tab:retained} is the spread rather than any
single score. RACE, PIQA, and CommonsenseQA retain 86--89\% of teacher performance,
while QASC and SciQ remain above 80\%. These results suggest that several forms of
plausibility, commonsense judgement, and relatively accessible knowledge remain
usable after conversion \cite{piqa,commonsenseqa}.

At the same time, the 77.1\% figure must be read as a \emph{selected interpretable
subset mean}, not as a claim that Cloe retains 77.1\% of all general intelligence.
Tasks for which both models collapse or for which the measurement instrument is
degenerate do not provide a defensible retention denominator.

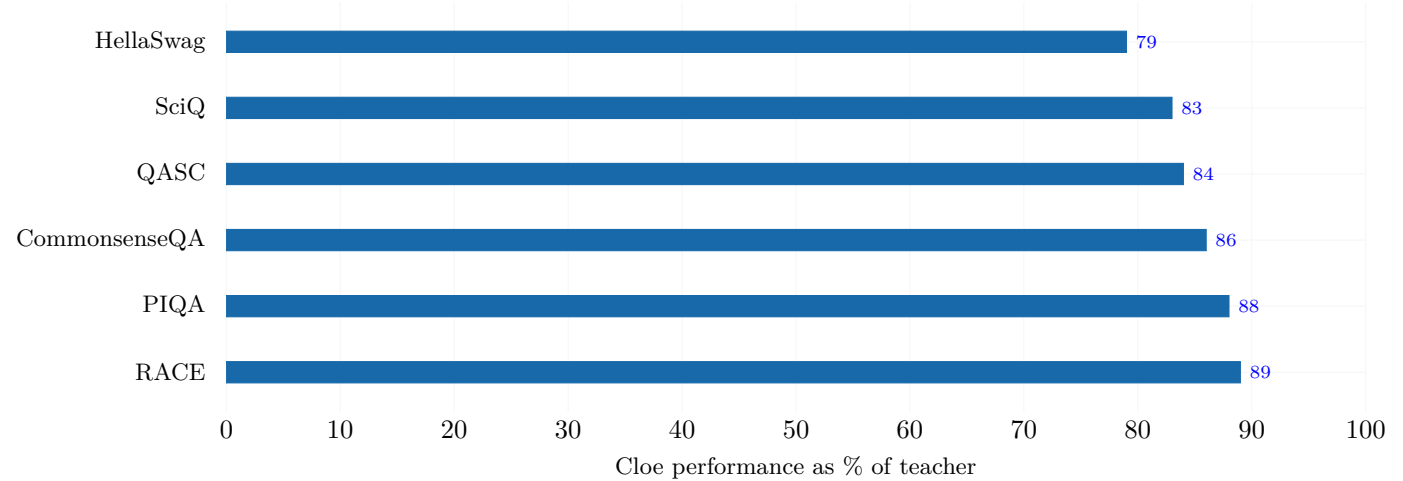
\begin{figure}[H]
\centering
\begin{tikzpicture}
\begin{axis}[
    width=0.96\textwidth,
    height=7.0cm,
    xbar,
    xmin=0,xmax=100,
    xlabel={Cloe performance as \% of teacher},
    symbolic y coords={RACE,PIQA,CommonsenseQA,QASC,SciQ,HellaSwag},
    ytick=data,
    nodes near coords,
    nodes near coords align={horizontal},
    every node near coord/.append style={font=\scriptsize\bfseries},
    bar width=8pt,
    axis line style={draw=none},
    tick style={draw=none},
    yticklabel style={font=\small},
    xlabel style={font=\small},
    grid=major,
    major grid style={draw=SoftGray},
    enlarge y limits=0.12,
]
\addplot+[fill=OneBitBlue,draw=OneBitBlue] coordinates
{(89,RACE) (88,PIQA) (86,CommonsenseQA) (84,QASC) (83,SciQ) (79,HellaSwag)};
\end{axis}
\end{tikzpicture}
\caption{Representative retained capabilities. The main visual message is the
non-uniform degradation profile: several commonsense and plausibility-oriented
tasks remain substantially above chance after ternarisation.}
\label{fig:retention}
\end{figure}

\begin{table}[H]
\centering
\caption{Capability boundaries and evaluation-limited tasks. ``Bar'' denotes
the baseline ternary model, while ``Over bar'' reports the absolute Cloe
improvement over that baseline. Retention is reported relative to the
full-precision teacher; unusually large retention values are shown without
positive highlighting when the teacher score is itself very low.}
\label{tab:boundaries}
\small
\setlength{\tabcolsep}{4pt}
\begin{tabularx}{\textwidth}{
    l
    X
    >{\centering\arraybackslash}p{1.35cm}
    >{\centering\arraybackslash}p{1.35cm}
    >{\centering\arraybackslash}p{1.35cm}
    >{\centering\arraybackslash}p{1.15cm}
    >{\centering\arraybackslash}p{1.25cm}
}
\toprule
\textbf{Task} &
\textbf{Capability} &
\textbf{Cloe} &
\textbf{Teacher} &
\textbf{Retention} &
\textbf{Bar} &
\textbf{Over bar} \\
\midrule

\texttt{winogrande}
& Coreference
& \textbf{53.7\%}
& 55.3\%
& \cellcolor{SoftGreen}\good{97\%}
& 50.0\%
& +3.7 \\

\texttt{gsm8k}
& Arithmetic reasoning
& \textbf{2.0\%}
& 4.0\%
& 50\%
& 0.0\%
& +2.0 \\

\texttt{copa}
& Causal commonsense
& \textbf{57.0\%}
& 64.0\%
& \cellcolor{SoftGreen}\good{89\%}
& 55.0\%
& +2.0 \\

\texttt{rte}
& Entailment
& \textbf{53.4\%}
& 52.7\%
& 101\%
& 52.7\%
& +0.7 \\

\texttt{boolq}
& Reading
& \textbf{62.7\%}
& 62.7\%
& 100\%
& 62.7\%
& +0.0 \\

\texttt{mrpc}
& Paraphrase
& \textbf{69.7\%}
& 69.7\%
& 100\%
& 69.7\%
& +0.0 \\

\texttt{cola}
& Grammaticality
& \textbf{65.7\%}
& 65.7\%
& 100\%
& 65.7\%
& +0.0 \\

\texttt{arc\_challenge}
& Knowledge-hard
& \textbf{27.1\%}
& 38.5\%
& 70\%
& 27.8\%
& -0.7 \\

\texttt{openbookqa}
& Knowledge-easy
& \textbf{26.7\%}
& 30.7\%
& 87\%
& 28.0\%
& -1.3 \\

\texttt{cb}
& Entailment
& \textbf{48.2\%}
& 8.9\%
& 540\%
& 50.0\%
& -1.8 \\

\texttt{qnli}
& Entailment
& \textbf{49.0\%}
& 49.0\%
& 100\%
& 51.0\%
& -2.0 \\

\texttt{wic}
& Word sense
& \textbf{48.7\%}
& 48.7\%
& 100\%
& 51.3\%
& -2.7 \\

\texttt{mmlu}
& Knowledge-hard
& \textbf{27.3\%}
& 27.0\%
& 101\%
& 31.7\%
& -4.3 \\

\texttt{medmcqa}
& Knowledge-expert
& \textbf{25.7\%}
& 27.0\%
& 95\%
& 31.3\%
& -5.7 \\

\texttt{wnli}
& Entailment
& \textbf{43.7\%}
& 43.7\%
& 100\%
& 56.3\%
& -12.7 \\

\texttt{wsc}
& Coreference
& \textbf{36.5\%}
& 36.5\%
& 100\%
& 63.5\%
& -26.9 \\

\texttt{truthfulqa}
& Truthfulness
& \textbf{28.3\%}
& 26.3\%
& 108\%
& 100.0\%
& -71.7 \\

\bottomrule
\end{tabularx}
\end{table}

% ============================================================
\section{Locating the Lost Information}

The behavioural results raise a deeper question. If some benchmarks fail, is the
underlying information still encoded in the hidden states but inaccessible to the
standard output head, or has the conversion actually removed the information?

\subsection{Linear-probe diagnostic}

A multinomial logistic-regression probe is trained on final-layer hidden states
from 8,000 MMLU auxiliary-training examples and evaluated on all 14,042 test
questions. The probe was itself validated on synthetic representations before
being applied to Cloe: it recovered 100\% accuracy when the answer was explicitly
linearly encoded and approximately chance when the signal was absent.

\begin{table}[H]
\centering
\caption{Representation-level probe results. Chance for four-way MMLU classification is 25\%.}
\label{tab:probe}
\small
\begin{tabular}{lccc}
\toprule
\textbf{Representation} & \textbf{Probe accuracy} & \textbf{Own answer accuracy} & \textbf{Effective rank}\\
\midrule
BF16 teacher & \cellcolor{SoftGreen}\good{43.76\%} & 44.72\% & 557.5\\
Cloe, post-SFT & 26.19\% & 24.31\% & \cellcolor{SoftBlue}\textbf{602.1}\\
Cloe, pre-SFT & 26.68\% & 22.97\% & 538.9\\
\midrule
\textit{Chance} & \textit{25.00\%} & \textit{25.00\%} & --\\
\bottomrule
\end{tabular}
\end{table}

The probe result is one of the strongest mechanistic observations in the study.
Cloe's 26.19\% probe accuracy is only about 1.2 percentage points above chance,
whereas the teacher reaches 43.76\%. Since the probe provides a relatively generous
linear readout of the hidden state, the result supports the interpretation that
specialist MMLU information is substantially absent from Cloe's representation,
rather than simply hidden behind a poorly calibrated output head.

\begin{figure}[H]
\centering
\begin{tikzpicture}
\begin{axis}[
    ybar,
    bar width=20pt,
    width=0.75\textwidth,
    height=6.0cm,
    ymin=0,ymax=50,
    ylabel={Linear-probe accuracy (\%)},
    symbolic x coords={Teacher,Cloe},
    xtick=data,
    nodes near coords,
    nodes near coords style={font=\small\bfseries},
    axis line style={draw=none},
    tick style={draw=none},
    grid=major,
    major grid style={draw=SoftGray},
    legend style={at={(0.5,-0.20)},anchor=north},
]
\addplot+[fill=OneBitBlue,draw=OneBitBlue] coordinates {(Teacher,43.76) (Cloe,26.19)};
\addplot+[sharp plot,no marks,draw=OneBitRed,very thick] coordinates {(Teacher,25) (Cloe,25)};
\legend{Probe accuracy,Chance}
\end{axis}
\end{tikzpicture}
\caption{MMLU information recoverable from final-layer representations. Cloe approaches
the 25\% four-way chance level while the teacher retains substantial recoverable signal.}
\label{fig:probe}
\end{figure}
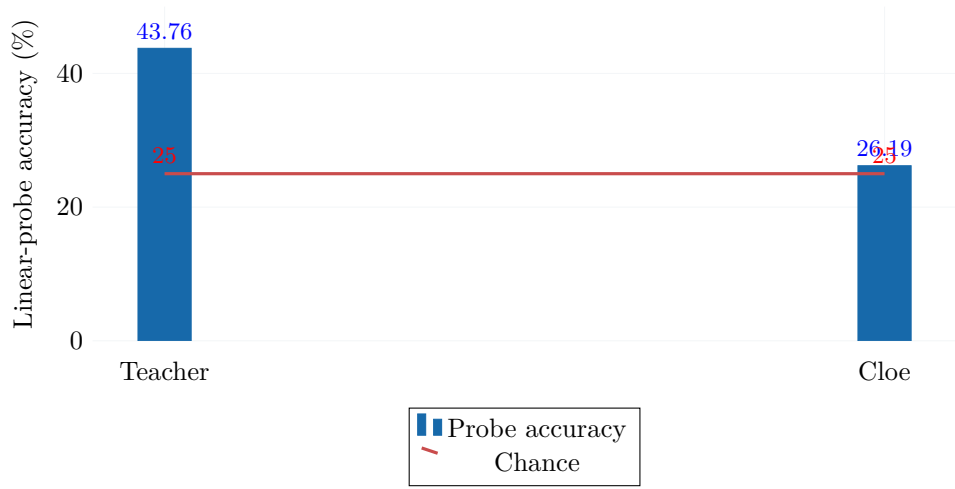

% ============================================================
\section{Representation Capacity Survives}

The probe result establishes that a particular class of specialist information is
lost. It does not, however, imply that the entire representation has collapsed.
To test this distinction, we measure effective rank of the hidden-state
representations.

\begin{table}[H]
\centering
\caption{Effective-rank comparison.}
\label{tab:rank}
\small
\begin{tabular}{lcc}
\toprule
\textbf{Model} & \textbf{Effective rank} & \textbf{Relative to teacher}\\
\midrule
BF16 teacher & 557.5 & 100.0\%\\
Cloe, pre-SFT & 538.9 & 96.7\%\\
Cloe, post-SFT & \cellcolor{SoftGreen}\textbf{602.1} & \cellcolor{SoftGreen}\textbf{108.0\%}\\
\bottomrule
\end{tabular}
\end{table}

The result is notable because the post-SFT Cloe representation has an effective
rank of 602.1 compared with 557.5 for the teacher. Thus the observed knowledge
loss cannot be reduced to a simple ``all information collapsed into a few
directions'' explanation. Cloe has substantial representational diversity even
while specific specialist information is poorly recoverable.

\begin{insightbox}
\textbf{Key distinction.}
The evidence supports a separation between \emph{representational capacity} and
\emph{stored knowledge}. Ternarisation can preserve a rich activation geometry
without preserving all of the fine-grained factual information that was encoded
in the original pretrained weights.
\end{insightbox}

% ============================================================
\section{Downstream Learnability}

The representation result becomes substantially more useful when combined with a
learning experiment. If the compressed model has lost broad factual knowledge but
retains useful representational structure, can it still acquire a new application
through fine-tuning?

We answer this using matched teacher controls: the teacher is fine-tuned under the
same downstream setup, so Cloe is compared with the teacher after both have been
adapted to the task.

\subsection{SST-2 sentiment classification}

Cloe begins at 61.9\% and reaches 89.8\% after fine-tuning. Relative to the matched
fine-tuned teacher, this corresponds to 95.6\% retention. The absolute gap is
reported alongside the retention number because a retention ratio without the
teacher's absolute score can exaggerate the apparent quality of a weak baseline.

The important observation is not simply that the score increased. It is that a
model whose specialist knowledge is substantially degraded can still reorganise
its parameters sufficiently to acquire a downstream discriminative task. The
measured learning gap is approximately 4.13 percentage points with an estimated
standard error of $\pm2.39$ points in the reported experiment.

\subsection{XSum summarisation}

The same pattern is observed on a generative task. Cloe improves from 11.1 to
19.0 ROUGE-L and reaches 79.4\% of the matched teacher \cite{rouge}. It also exceeds the
LEAD-1 copy/input-order baseline by approximately 2.4$\times$, supporting the
interpretation that the improvement reflects actual summarisation learning rather
than simple copying \cite{xsum}.

\begin{table}[H]
\centering
\caption{Matched downstream learning results.}
\label{tab:finetune}
\small
\begin{tabular}{lccc}
\toprule
\textbf{Task} & \textbf{Cloe before} & \textbf{Cloe after} & \textbf{Teacher retention}\\
\midrule
SST-2 & 61.9\% & \cellcolor{SoftGreen}\textbf{89.8\%} & \cellcolor{SoftGreen}\textbf{95.6\%}\\
XSum & 11.1 ROUGE-L & \cellcolor{SoftGreen}\textbf{19.0 ROUGE-L} & \cellcolor{SoftBlue}\textbf{79.4\%}\\
\bottomrule
\end{tabular}
\end{table}

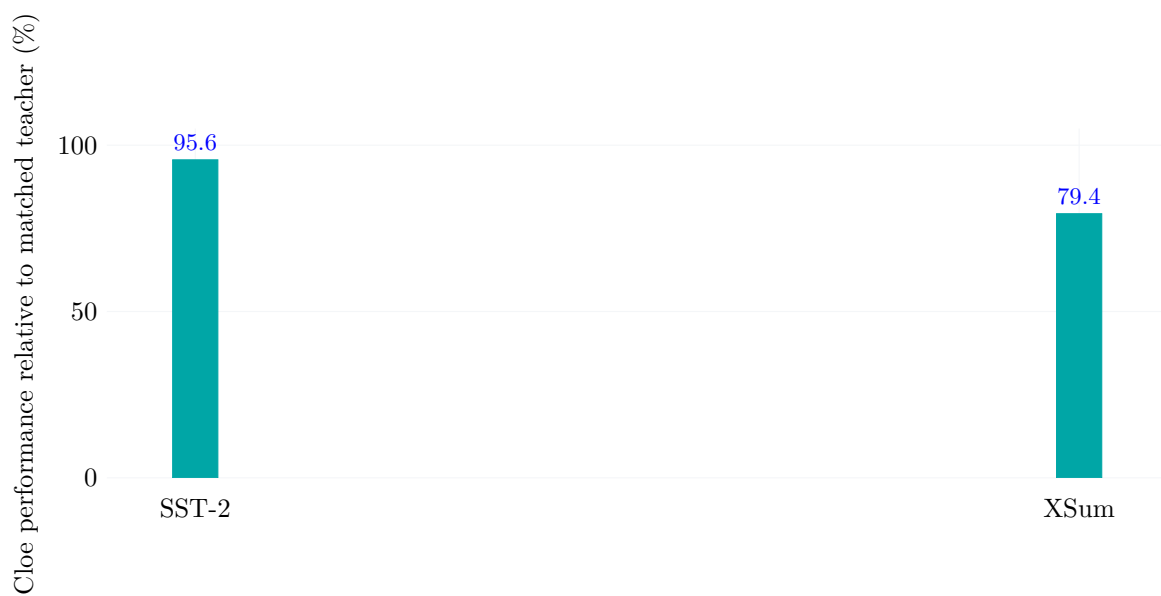
\begin{figure}[H]
\centering
\begin{tikzpicture}
\begin{axis}[
    ybar,
    bar width=17pt,
    width=0.90\textwidth,
    height=6.2cm,
    ymin=0,ymax=105,
    ylabel={Cloe performance relative to matched teacher (\%)},
    symbolic x coords={SST-2,XSum},
    xtick=data,
    nodes near coords,
    nodes near coords style={font=\small\bfseries},
    axis line style={draw=none},
    tick style={draw=none},
    grid=major,
    major grid style={draw=SoftGray},
]
\addplot+[fill=OneBitCyan,draw=OneBitCyan] coordinates {(SST-2,95.6) (XSum,79.4)};
\end{axis}
\end{tikzpicture}
\caption{Downstream learnability expressed relative to identically fine-tuned teacher
controls. The absolute task scores remain important and are reported in \cref{tab:finetune}.}
\label{fig:learnability}
\end{figure}

\begin{insightbox}
\textbf{Why this matters.}
The combination of the probe and fine-tuning results gives the paper its strongest
conceptual separation: Cloe does not preserve all of the teacher's pretrained
knowledge, but it retains enough representational structure to learn useful
application-specific behaviour. This is the basis for viewing extreme
quantisation as potentially useful for specialised deployment rather than only
as an attempt to reproduce a general-purpose teacher.
\end{insightbox}

% ============================================================
\section{Capability Profile: What Changes and What Survives?}

The results can be condensed into a capability map. This table is intentionally
more important than a long narrative: the reader should be able to understand the
operating envelope of the model before reading the interpretation below it.

\begin{table}[H]
\centering
\caption{Capability-stratified outcome profile of ternarisation.}
\label{tab:profile}
\small
\begin{tabularx}{\textwidth}{p{3.0cm}p{2.1cm}p{3.0cm}Y}
\toprule
\textbf{Property} & \textbf{Direction} & \textbf{Evidence} & \textbf{Interpretation}\\
\midrule
Specialist factual knowledge & \cellcolor{SoftRed}\bad{$\downarrow\downarrow\downarrow$}
& MMLU + linear probe & Much of the teacher's recoverable specialist information is lost.\\

Arithmetic reasoning & \cellcolor{SoftRed}\bad{$\downarrow\downarrow\downarrow$}
& GSM8K & Not a reliable general arithmetic model after conversion.\\

Language modelling & \cellcolor{SoftRed}\bad{$\downarrow\downarrow$}
& LAMBADA / WikiText-2 & Token-level modelling quality degrades substantially.\\

Commonsense judgement & \cellcolor{SoftGreen}\good{$\rightarrow$}
& 10-task benchmark subset & Several plausibility-oriented capabilities remain.\\

Representation diversity & \cellcolor{SoftGreen}\good{$\rightarrow$}
& Effective rank & No evidence of broad representational collapse.\\

Downstream learnability & \cellcolor{SoftGreen}\good{$\rightarrow$}
& SST-2 / XSum & Task-specific adaptation remains effective.\\

Packed storage & \cellcolor{SoftBlue}\textbf{$\uparrow\uparrow$}
& Deployment benchmark & Packed artifact is substantially smaller than latent representation.\\

Inference throughput & \cellcolor{SoftBlue}\textbf{$\uparrow$}
& Packed benchmark & Prefill and decode are faster in the measured packed-vs-latent comparison.\\
\bottomrule
\end{tabularx}
\end{table}

The table suggests that the conversion should not be interpreted as a single scalar
loss of intelligence. Different properties occupy different layers of the system.
The most severe losses appear in specialist factual retrieval, arithmetic, and
language modelling. In contrast, the model retains useful performance on a subset
of commonsense and plausibility tasks, shows substantial hidden-state diversity,
and can be adapted to new downstream objectives.

This stratification is the central empirical result of the study. It also changes
how a compressed model should be evaluated. If the intended product only requires a
narrow task after supervised adaptation, the relevant question is not whether the
compressed model retains all of the teacher's world knowledge. The relevant question
is whether the compressed substrate retains enough capacity to learn the required
task at the target deployment cost.

% ============================================================
\section{Deployment and Efficiency Characterisation}
All deployment measurements were collected on a fixed workstation
equipped with an AMD Ryzen 9 7900X CPU, 32 GB system memory, and an
NVIDIA GeForce RTX 5070 GPU with 11.94 GiB of VRAM. The experiments were conducted under
Windows 10 Pro using Python 3.10.11, PyTorch
2.12.0.dev20260408+cu128, CUDA 12.8, and Transformers 4.51.0.
PyTorch was configured with 12 CPU threads, and no explicit
\texttt{CUDA\_VISIBLE\_DEVICES}, \texttt{OMP\_NUM\_THREADS}, or
\texttt{MKL\_NUM\_THREADS} overrides were present in the environment.
Both the latent and packed representations were evaluated under the
same software and hardware configuration.

\begin{table}[h]
\centering
\caption{Hardware and software environment used for deployment measurements.}
\label{tab:environment}
\begin{tabular}{ll}
\toprule
\textbf{Component} & \textbf{Configuration} \\
\midrule
CPU & AMD Ryzen 9 7900X (12 cores, 24 threads) \\
System memory & 32 GB RAM \\
GPU & NVIDIA GeForce RTX 5070 \\
GPU VRAM& 12,227 MiB (~11.94 GiB) \\
Operating system & Windows 10 Pro (64-bit) \\
Python & 3.10.11 \\
PyTorch & 2.12.0.dev20260408+cu128 \\
CUDA & 12.8 \\
Transformers & 4.51.0 \\
PyTorch CPU threads & 12 \\
\bottomrule
\end{tabular}
\end{table}

The capability study establishes what the model retains. We now examine the
engineering consequence of packing the ternary representation.

The current deployment comparison is between a \emph{packed} representation and
the measured \emph{latent} representation. Because the exact runtime semantics of
the latent representation differ from those of a conventional FP/BF16 baseline,
we do not use this comparison to claim a universal speedup over all full-precision
implementations. Instead, it is reported as a representation-level deployment
measurement.

\begin{table}[H]
\centering
\caption{Packed versus latent deployment measurements from the current benchmark.}
\label{tab:deployment}
\small
\begin{tabular}{lrrrr}
\toprule
\textbf{Metric} & \textbf{Packed} & \textbf{Latent} & \textbf{Relative change} &\textbf{Interpretation}\\
\midrule
Disk footprint & \cellcolor{SoftGreen}\textbf{634.65 MB}
    & 1505.60 MB & \cellcolor{SoftGreen}\textbf{2.37$\times$ smaller} & Major storage reduction\\
Weight memory & 1578.36 MiB & 1460.59 MiB & +8.1\% &Runtime memory overhead\\
Peak memory & 1708.98 MiB & 1591.22 MiB & +7.4\% & Modest peak-memory increase\\
Load time & 2.858 s & \cellcolor{SoftGreen}\textbf{1.570 s} & 1.82$\times$ slower & Initialization penalty\\
Prefill & \cellcolor{SoftGreen}\textbf{1806.76 tok/s}
    & 1477.03 tok/s & \cellcolor{SoftGreen}\textbf{+22.3\%} & Faster steady-state processing\\
Decode & \cellcolor{SoftGreen}\textbf{7.634 tok/s}
    & 6.145 tok/s & \cellcolor{SoftGreen}\textbf{+24.2\%} & Faster token generation\\
\bottomrule
\end{tabular}
\end{table}

Three observations stand out from \cref{tab:deployment}. First, the packed artifact
is dramatically smaller on disk: approximately 634.7 MB versus 1.50 GB for the
latent representation. This is the clearest storage benefit in the current
measurement. Second, the packed path shows approximately 22.3\% higher prefill
throughput and 24.2\% higher decode throughput in this particular comparison.
Third, storage compression does not translate one-to-one into runtime memory:
peak memory is approximately 1.67 GiB for the packed path and slightly higher than
the latent measurement.

The deployment measurements reveal a trade-off rather than a uniformly dominant representation. The packed model substantially reduces persistent storage requirements and provides higher steady-state inference throughput, while incurring modest runtime-memory overhead and a longer initialization time. Importantly, the storage, loading, and inference measurements correspond to distinct stages of deployment and therefore need not move in the same direction. This last point is important for product engineering. A compact model file and a
low-memory runtime are related but distinct optimisation targets. The deployment
stack may need additional buffers, unpacking structures, kernels, or other runtime
state. Consequently, the paper reports both disk footprint and peak memory rather
than using one as a proxy for the other.

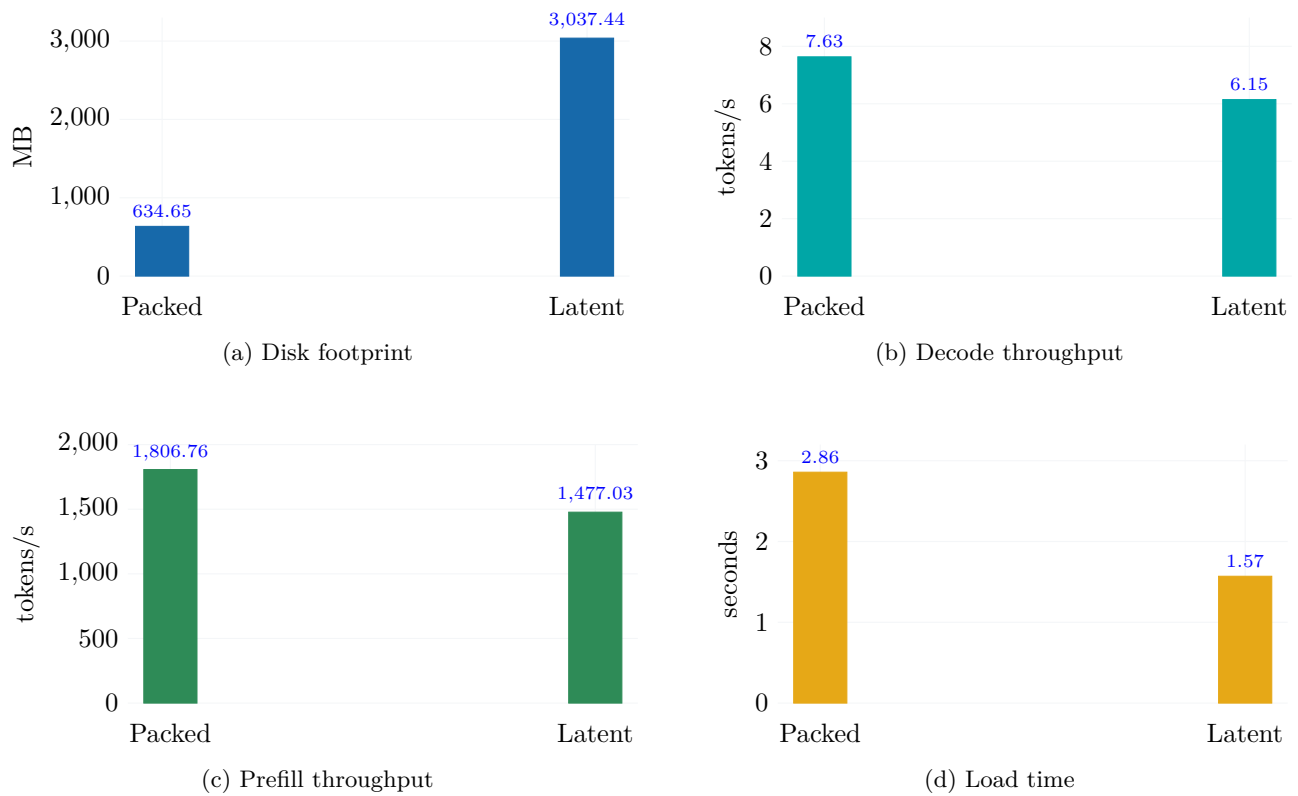
\begin{figure}[H]
\centering
\begin{subfigure}{0.48\textwidth}
\centering
\begin{tikzpicture}
\begin{axis}[
    ybar,
    bar width=20pt,
    width=\linewidth,
    height=5.0cm,
    ymin=0,ymax=3300,
    ylabel={MB},
    symbolic x coords={Packed,Latent},
    xtick=data,
    nodes near coords,
    every node near coord/.style={font=\scriptsize\bfseries},
    axis line style={draw=none},
    tick style={draw=none},
    grid=major,
    major grid style={draw=SoftGray},
]
\addplot+[fill=OneBitBlue,draw=OneBitBlue]
coordinates {(Packed,634.65) (Latent,3037.44)};
\end{axis}
\end{tikzpicture}
\caption{Disk footprint}
\end{subfigure}
\hfill
\begin{subfigure}{0.48\textwidth}
\centering
\begin{tikzpicture}
\begin{axis}[
    ybar,
    bar width=20pt,
    width=\linewidth,
    height=5.0cm,
    ymin=0,ymax=9,
    ylabel={tokens/s},
    symbolic x coords={Packed,Latent},
    xtick=data,
    nodes near coords,
    nodes near coords style={font=\scriptsize\bfseries},
    axis line style={draw=none},
    tick style={draw=none},
    grid=major,
    major grid style={draw=SoftGray},
]
\addplot+[fill=OneBitCyan,draw=OneBitCyan]
coordinates {(Packed,7.634) (Latent,6.145)};
\end{axis}
\end{tikzpicture}
\caption{Decode throughput}
\end{subfigure}

\vspace{7mm}

\begin{subfigure}{0.48\textwidth}
\centering
\begin{tikzpicture}
\begin{axis}[
    ybar,
    bar width=20pt,
    width=\linewidth,
    height=5.0cm,
    ymin=0,ymax=2000,
    ylabel={tokens/s},
    symbolic x coords={Packed,Latent},
    xtick=data,
    nodes near coords,
    nodes near coords style={font=\scriptsize\bfseries},
    axis line style={draw=none},
    tick style={draw=none},
    grid=major,
    major grid style={draw=SoftGray},
]
\addplot+[fill=OneBitGreen,draw=OneBitGreen]
coordinates {(Packed,1806.76) (Latent,1477.03)};
\end{axis}
\end{tikzpicture}
\caption{Prefill throughput}
\end{subfigure}
\hfill
\begin{subfigure}{0.48\textwidth}
\centering
\begin{tikzpicture}
\begin{axis}[
    ybar,
    bar width=20pt,
    width=\linewidth,
    height=5.0cm,
    ymin=0,ymax=3.2,
    ylabel={seconds},
    symbolic x coords={Packed,Latent},
    xtick=data,
    nodes near coords,
    nodes near coords style={font=\scriptsize\bfseries},
    axis line style={draw=none},
    tick style={draw=none},
    grid=major,
    major grid style={draw=SoftGray},
]
\addplot+[fill=OneBitGold,draw=OneBitGold]
coordinates {(Packed,2.858) (Latent,1.570)};
\end{axis}
\end{tikzpicture}
\caption{Load time}
\end{subfigure}
\caption{Current packed-versus-latent deployment measurements. The strongest
observed benefit is storage footprint, while throughput also improves in the
measured execution path. Runtime memory and loading remain separate engineering
considerations.}
\label{fig:deployment}
\end{figure}

% ============================================================
\section{Failure Modes as Operating-Boundary Conditions}

The study includes negative results because a commercially useful technical paper
must identify where the system should \emph{not} be used. These results are therefore
presented as operating boundaries rather than as the central narrative.

\begin{table}[H]
\centering
\caption{Selected limitations and what they imply for deployment.}
\label{tab:limitations}
\small
\begin{tabularx}{\textwidth}{p{3.0cm}p{3.0cm}Y}
\toprule
\textbf{Consideration} & \textbf{Current evidence} & \textbf{Engineering implication}\\
\midrule
Specialist knowledge & MMLU / ARC-C / MedMCQA near chance &
Do not treat Cloe as a drop-in replacement for a broad knowledge model.\\

Arithmetic & GSM8K $\approx$ 2.0\% &
Applications requiring reliable symbolic arithmetic need dedicated mechanisms.\\

Truthfulness & TruthfulQA 28.3\%; high unanswerable-question confabulation &
Safety-sensitive factual deployment requires task-specific safeguards.\\

Language modelling & LAMBADA perplexity 12.1$\times$ worse; WikiText-2 3.1$\times$ worse &
General free-form generation quality is materially degraded.\\

Whole-checkpoint compression & Embeddings remain BF16; overall checkpoint $\approx$6.47 bits/parameter &
The 1.58-bit figure should be reserved for the ternary weight representation.\\

Runtime memory & Packed peak $\approx$1.67 GiB &
Disk compression should not be interpreted as equal runtime-memory compression.\\

Fine-tuning uncertainty & Single reported seed &
Additional seeds are needed for production-grade variance estimates.\\

Quantiser comparison & No PTQ control in the current study &
The study does not claim QAT is superior to every alternative conversion recipe.\\
\bottomrule
\end{tabularx}
\end{table}

The limitations do not invalidate the main capability-stratification result. Instead,
they define its scope. The current Cloe checkpoint should not be presented as a
general-purpose replacement for the teacher. The stronger and more defensible
interpretation is that extreme ternarisation produces a model with a distinctive
capability profile: broad pretrained knowledge is substantially damaged, but
representational diversity and task-specific learnability remain useful.

% ============================================================
\section{Application and Product Implications}

The technical findings suggest a deployment paradigm different from simply trying
to preserve every capability of a large general-purpose model.

\subsection{From knowledge preservation to task-specific adaptation}

The combined results can be viewed as a three-stage transformation:

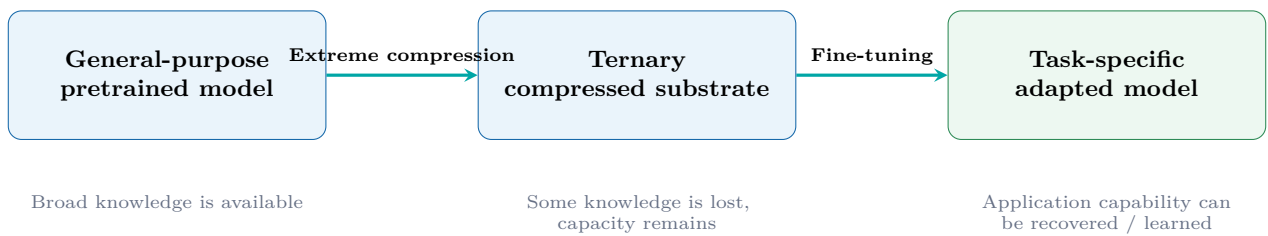
\begin{figure}[H]
\centering
\begin{tikzpicture}[
    node distance=11mm and 20mm,
    stage/.style={
        rounded corners=5pt,
        minimum width=42mm,
        minimum height=17mm,
        align=center,
        font=\small\bfseries,
        draw=OneBitBlue,
        fill=SoftBlue
    },
    arrow/.style={
        ->,
        >=stealth,
        very thick,
        draw=OneBitCyan
    },
    note/.style={
        font=\scriptsize\color{MidGray},
        align=center,
        text width=42mm
    }
]
\node[stage] (general) {General-purpose\\pretrained model};

\node[stage,right=of general] (compressed)
    {Ternary\\compressed substrate};

\node[
    stage,
    right=of compressed,
    fill=SoftGreen,
    draw=OneBitGreen
] (special)
    {Task-specific\\adapted model};

\draw[arrow]
    (general)--node[above,font=\scriptsize\bfseries]
    {Extreme compression}(compressed);

\draw[arrow]
    (compressed)--node[above,font=\scriptsize\bfseries]
    {Fine-tuning}(special);

\node[note,below=6mm of general]
    {Broad knowledge is available};

\node[note,below=6mm of compressed]
    {Some knowledge is lost,\\capacity remains};

\node[note,below=6mm of special]
    {Application capability can\\be recovered / learned};

\end{tikzpicture}
\caption{Conceptual deployment thesis suggested by the combined experiments.}
\label{fig:product}
\end{figure}

The proposed interpretation is therefore not that every application should replace
a general-purpose model with Cloe. Instead, applications with narrow or well-defined
objectives may value a compact substrate whose task-specific capability can be
learned after compression.

\begin{table}[H]
\centering
\caption{Potential application mapping. These are deployment hypotheses, not claims
that every scenario has been experimentally validated in this study.}
\label{tab:applications}
\small
\begin{tabularx}{\textwidth}{p{3.0cm}p{3.4cm}Y}
\toprule
\textbf{Scenario} & \textbf{Why compression matters} & \textbf{Evidence / condition}\\
\midrule
Edge / constrained inference & Smaller model artifact can reduce storage and transfer burden &
Current packed artifact is 634.7 MB.\\

Private or local assistants & Smaller local model can reduce dependence on remote inference &
Requires further end-to-end device validation.\\

Task-specific classifiers & General knowledge is less important than adaptation &
SST-2 reaches 95.6\% matched-teacher retention.\\

Specialised summarisation & Task learnability matters more than broad pretrained knowledge &
XSum reaches 79.4\% teacher retention after adaptation.\\

Cost-sensitive inference & Throughput and representation size can affect serving economics &
Current packed path shows +22.3\% prefill and +24.2\% decode versus latent measurement.\\

General-purpose factual assistant & Broad pretrained knowledge is essential &
\bad{Not recommended from current evidence.}\\
\bottomrule
\end{tabularx}
\end{table}

The market-facing implication should therefore be framed around \emph{deployment
economics and specialization}, not around an unsupported claim that ternary
conversion preserves general intelligence. The current results provide evidence for
compact storage, measurable inference benefits in the tested packed path, and
strong downstream adaptation on selected tasks. They do not establish a universal
advantage across hardware, model scales, or workloads.

% ============================================================
\section{Discussion}

\subsection{A capability-stratified view of quantisation}

The most useful conceptual outcome of the study is a separation of three properties
that are often conflated:

\begin{enumerate}[leftmargin=1.5em,itemsep=2pt]
    \item \textbf{Stored knowledge:} substantially reduced, particularly for
    specialist factual information.
    \item \textbf{Representational capacity:} not obviously destroyed, as indicated
    by effective rank.
    \item \textbf{Learnability:} remains substantial on the tested downstream tasks.
\end{enumerate}

This separation provides a more informative description than calling the conversion
simply ``successful'' or ``failed.'' It explains why a model can simultaneously
perform poorly on broad knowledge benchmarks and still reach near-teacher
performance after supervised adaptation on a targeted task.

\subsection{Why the evaluation methodology is part of the contribution}

The initial MMLU result demonstrates that low-bit models can expose failure modes
that conventional evaluation pipelines are not designed to distinguish. Answer-letter
bias can convert a degenerate output policy into an apparently meaningful accuracy
number. Likewise, retention ratios can become misleading when the teacher itself
collapses on a task.

The practical lesson is that low-bit model evaluation should combine behavioural
benchmarks with representation-level and learning-level diagnostics. A useful
evaluation stack therefore contains:

\[
\boxed{
\text{Benchmark}
\rightarrow
\text{Guard}
\rightarrow
\text{Representation probe}
\rightarrow
\text{Matched adaptation}
}
\]

This is particularly important when the research goal is to understand the
\emph{deployment envelope} rather than merely to rank checkpoints.

\subsection{What the results do not establish}

The study does not establish that QAT conversion is universally superior to PTQ,
that Cloe is a general-purpose replacement for Qwen3.5-0.8B, or that the current
packed benchmark translates directly into lower cost on every target device.
Similarly, the 1.5821 bits/weight entropy describes the ternary weight tensors,
whereas the complete checkpoint remains larger because the embedding/LM-head
structure is not ternarised.

The present study does not include a recovery-budget ablation. Consequently, the observed capability gap cannot be uniquely decomposed into representational loss caused by ternarisation and capability that remains unrecovered because of limited QAT. A controlled sweep over QAT budgets (e.g., 20M, 72.4M, and 200M tokens) would establish the capability–compute recovery curve and provide a stronger estimate of the irreducible effect of ternarisation.

% ============================================================
\section{Conclusion}

This study investigates what happens when a pretrained small language model is
converted into a ternary representation using quantisation-aware training. The
answer is not uniform degradation. Instead, the conversion produces a structured
capability profile.

The most significant loss is specialist pretrained knowledge: the linear probe
falls from 43.76\% for the BF16 teacher to 26.19\% for Cloe, close to the 25\%
chance level. At the same time, Cloe retains meaningful performance across a subset
of commonsense and plausibility-oriented benchmarks, with a mean 77.1\% teacher
retention on ten demonstrably interpretable tasks. Effective rank remains high,
and downstream adaptation is particularly encouraging: Cloe reaches 95.6\% of the
matched teacher on SST-2 and 79.4\% on XSum. Because all reported post-quantization results correspond to a single 72.4M token recovery budget, these measurements characterize the outcome of this recovery regime rather than an irreducible capability limit of the ternary representation.

The engineering measurements add a second dimension. The current packed artifact
occupies approximately 634.7 MB on disk and shows higher prefill and decode
throughput than the measured latent representation. Runtime memory, however, does
not fall in direct proportion to disk footprint, demonstrating why storage,
runtime memory, and throughput should be measured separately.

Taken together, the results support a practical thesis:

\begin{tcolorbox}[
    colback=SoftBlue,
    borderline west={4pt}{0pt}{OneBitBlue},
    width=\textwidth,
    halign=center
]
\large\bfseries
Extreme quantisation can destroy broad pretrained knowledge without destroying
the ability to represent and learn useful application-specific behaviour.
\end{tcolorbox}

% ============================================================
\newpage

% ============================================================
% REFERENCES
% ============================================================


\begin{thebibliography}{99}

% -------------------- Base model ----------------------------


\bibitem{qwen3}
A. Yang et al.
\newblock Qwen3 Technical Report.
\newblock arXiv:2505.09388, 2025.
\newblock \url{https://arxiv.org/abs/2505.09388}.

% -------------------- Ternary / 1.58-bit --------------------

\bibitem{bitnet}
H. Wang, S. Ma, L. Dong, S. Huang, H. Wang,
L. Ma, F. Yang, R. Wang, Y. Wu, and F. Wei.
\newblock BitNet: Scaling 1-bit Transformers for Large Language Models.
\newblock arXiv:2310.11453, 2023.
\newblock \url{https://arxiv.org/abs/2310.11453}.

\bibitem{bitnetb158}
S. Ma, H. Wang, L. Ma, L. Wang, W. Wang,
S. Huang, L. Dong, R. Wang, J. Xue, and F. Wei.
\newblock The Era of 1-bit LLMs: All Large Language Models are in 1.58 Bits.
\newblock arXiv:2402.17764, 2024.
\newblock \url{https://arxiv.org/abs/2402.17764}.


% -------------------- Quantization --------------------------

\bibitem{gptq}
E. Frantar, S. Ashkboos, T. Hoefler, and D. Alistarh.
\newblock GPTQ: Accurate Post-Training Quantization for Generative
Pre-trained Transformers.
\newblock In \emph{International Conference on Learning Representations
(ICLR)}, 2023.
\newblock arXiv:2210.17323.
\newblock \url{https://arxiv.org/abs/2210.17323}.

\bibitem{lowbitundertrained}
X. Ouyang, T. Ge, T. Hartvigsen, Z. Zhang, H. Mi, and D. Yu.
\newblock Low-Bit Quantization Favors Undertrained LLMs:
Scaling Laws for Quantized LLMs with 100T Training Tokens.
\newblock In \emph{Proceedings of the 63rd Annual Meeting of the
Association for Computational Linguistics}, 2025.
\newblock arXiv:2411.17691.
\newblock \url{https://arxiv.org/abs/2411.17691}.

\bibitem{distillation}
G. Hinton, O. Vinyals, and J. Dean.
\newblock Distilling the Knowledge in a Neural Network.
\newblock arXiv:1503.02531, 2015.
\newblock \url{https://arxiv.org/abs/1503.02531}.

% -------------------- Contemporary ternary work --------------

\bibitem{catq}
S. Wang, C. Li, Y. Kang, J. Fan, and A. Yao.
\newblock CAT-Q: Cost-efficient and Accurate Ternary Quantization for LLMs.
\newblock In \emph{International Conference on Machine Learning (ICML)},
2026.
\newblock arXiv:2606.26650.
\newblock \url{https://arxiv.org/abs/2606.26650}.

\bibitem{twla}
Z. Zhao, Z. Xu, Z. Chen, X. Hu, Z. Jiang, and D. Yang.
\newblock TWLA: Achieving Ternary Weights and Low-Bit Activations for
LLMs via Post-Training Quantization.
\newblock arXiv:2606.13054, 2026.
\newblock \url{https://arxiv.org/abs/2606.13054}.

\bibitem{ptqtp}
H. Xiao, R. Yang, Q. Yang, W. Xu, Z. Li,
Y. Su, Z. Liu, H. Yang, and N. Wong.
\newblock PTQTP: Post-Training Quantization to Trit-Planes for Large
Language Models.
\newblock arXiv:2509.16989, 2025.
\newblock \url{https://arxiv.org/abs/2509.16989}.

\bibitem{ayot}
S. Wang, C. Li, Y. Kang, J. Fan, and A. Yao.
\newblock Attend to Your Own Thoughts: Breaking the Barrier for
Post-Training Quantization of Reasoning LLMs through the Lens of
1.58-Bit Quantization.
\newblock arXiv:2608.01078, 2026.
\newblock \url{https://arxiv.org/abs/2608.01078}.

\bibitem{ternarymamba}
R. Ganesaraja, S. D. Panse, and S. N.
\newblock Ternary Mamba: Grouped Quantization-Aware Training of
W1.58A16 State Space Models.
\newblock arXiv:2606.18114, 2026.
\newblock \url{https://arxiv.org/abs/2606.18114}.

\bibitem{falconedge}
Falcon-LLM Team.
\newblock Falcon-E, a Series of Powerful, Universal and Fine-tunable
1.58bit Language Models.
\newblock Technical report / project publication, 2025.
\newblock \url{https://falcon-lm.github.io/blog/falcon-edge/}.


\bibitem{qwen35}
Qwen Team.
\newblock Qwen3.5: Towards Native Multimodal Agents.
\newblock 2026.
\newblock \url{https://qwen.ai/blog?id=qwen3.5}.

\bibitem{ste}
Y. Bengio, N. L\'{e}onard, and A. Courville.
\newblock Estimating or Propagating Gradients Through Stochastic Neurons
for Conditional Computation.
\newblock arXiv:1308.3432, 2013.
\newblock \url{https://arxiv.org/abs/1308.3432}.


\bibitem{rmsnorm158}
C. Steinmetz, G. Childress, A. Herbst, G. Jones,
J. Singh, E. Vang, and K. Weinstock.
\newblock An Extra RMSNorm is All You Need for Fine Tuning to 1.58 Bits.
\newblock arXiv:2505.08823, 2025.
\newblock \url{https://arxiv.org/abs/2505.08823}.

% -------------------- Evaluation -----------------------------

\bibitem{mmlu}
D. Hendrycks, C. Burns, S. Basart, A. Zou,
M. Mazeika, D. Song, and J. Steinhardt.
\newblock Measuring Massive Multitask Language Understanding.
\newblock In \emph{International Conference on Learning Representations
(ICLR)}, 2021.
\newblock arXiv:2009.03300.
\newblock \url{https://arxiv.org/abs/2009.03300}.

\bibitem{arc}
P. Clark, I. Cowhey, O. Etzioni, T. Khot,
A. Sabharwal, C. Schoenick, and O. Tafjord.
\newblock Think You Have Solved Question Answering? Try ARC, the AI2
Reasoning Challenge.
\newblock arXiv:1803.05457, 2018.
\newblock \url{https://arxiv.org/abs/1803.05457}.

\bibitem{openbookqa}
T. Mihaylov, P. Clark, T. Khot, and A. Sabharwal.
\newblock Can a Suit of Armor Conduct Electricity? A New Dataset for
Open Book Question Answering.
\newblock In \emph{Proceedings of EMNLP}, 2018.
\newblock \url{https://aclanthology.org/D18-1260/}.

\bibitem{sciq}
J. Welbl, N. F. Liu, and M. Gardner.
\newblock Crowdsourcing Multiple Choice Science Questions.
\newblock arXiv:1707.06209, 2017.
\newblock \url{https://arxiv.org/abs/1707.06209}.

\bibitem{qasc}
T. Khot, P. Clark, M. Guerquin, P. Jansen, and A. Sabharwal.
\newblock QASC: A Dataset for Question Answering via Sentence Composition.
\newblock In \emph{Proceedings of the AAAI Conference on Artificial
Intelligence}, 2020.
\newblock arXiv:1910.11473.
\newblock \url{https://arxiv.org/abs/1910.11473}.

\bibitem{hellaswag}
R. Zellers, A. Holtzman, Y. Bisk, A. Farhadi, and Y. Choi.
\newblock HellaSwag: Can a Machine Really Finish Your Sentence?
\newblock In \emph{ACL}, 2019.
\newblock arXiv:1905.07830.
\newblock \url{https://arxiv.org/abs/1905.07830}.

\bibitem{piqa}
Y. Bisk, R. Zellers, R. Le Bras, J. Gao, and Y. Choi.
\newblock PIQA: Reasoning about Physical Commonsense in Natural Language.
\newblock In \emph{AAAI}, 2020.
\newblock arXiv:1911.11641.
\newblock \url{https://arxiv.org/abs/1911.11641}.

\bibitem{commonsenseqa}
A. Talmor, J. Herzig, N. Lourie, and J. Berant.
\newblock CommonsenseQA: A Question Answering Challenge Targeting
Commonsense Knowledge.
\newblock In \emph{NAACL}, 2019.
\newblock arXiv:1811.00937.
\newblock \url{https://arxiv.org/abs/1811.00937}.

\bibitem{winogrande}
K. Sakaguchi, R. Le Bras, C. Bhagavatula, and Y. Choi.
\newblock WinoGrande: An Adversarial Winograd Schema Challenge at Scale.
\newblock In \emph{AAAI}, 2020.
\newblock arXiv:1907.10641.
\newblock \url{https://arxiv.org/abs/1907.10641}.

\bibitem{truthfulqa}
S. Lin, J. Hilton, and O. Evans.
\newblock TruthfulQA: Measuring How Models Mimic Human Falsehoods.
\newblock In \emph{ACL}, 2022.
\newblock arXiv:2109.07958.
\newblock \url{https://arxiv.org/abs/2109.07958}.

% -------------------- Benchmark suites ----------------------

\bibitem{glue}
A. Wang, A. Singh, J. Michael, F. Hill,
O. Levy, and S. R. Bowman.
\newblock GLUE: A Multi-Task Benchmark and Analysis Platform for
Natural Language Understanding.
\newblock In \emph{ICLR}, 2019.
\newblock arXiv:1804.07461.
\newblock \url{https://arxiv.org/abs/1804.07461}.

\bibitem{superglue}
A. Wang, Y. Pruksachatkun, N. Nangia,
A. Singh, J. Michael, F. Hill, O. Levy, and S. R. Bowman.
\newblock SuperGLUE: A Stickier Benchmark for General-Purpose Language
Understanding Systems.
\newblock In \emph{NeurIPS}, 2019.
\newblock arXiv:1905.00537.
\newblock \url{https://arxiv.org/abs/1905.00537}.

% -------------------- Downstream tasks -----------------------


\bibitem{rouge}
C.-Y. Lin.
\newblock ROUGE: A Package for Automatic Evaluation of Summaries.
\newblock In \emph{Text Summarization Branches Out}, ACL, 2004.

\bibitem{xsum}
S. Narayan, S. B. Cohen, and M. Lapata.
\newblock Don't Give Me the Details, Just the Summary! Topic-Aware
Convolutional Neural Networks for Extreme Summarization.
\newblock In \emph{EMNLP}, 2018.
\newblock arXiv:1808.08745.
\newblock \url{https://arxiv.org/abs/1808.08745}.


\end{thebibliography}
\end{document}